\documentclass[11pt,twocolumn]{article}

\usepackage[letterpaper,top=0.75in,bottom=1.0in,left=0.75in,right=0.75in,columnsep=0.25in]{geometry}
\usepackage{times}
\usepackage{latexsym}
\usepackage{microtype}
\usepackage{inconsolata}
\usepackage{booktabs}
\usepackage{tabularx}
\usepackage{multirow}
\usepackage{array}
\usepackage{enumitem}
\usepackage{xcolor}
\usepackage{amsmath}
\usepackage{graphicx}
\usepackage{tikz}
\usetikzlibrary{arrows.meta,positioning,fit,shapes.geometric,shapes.misc}
\usepackage[numbers,sort&compress]{natbib}
\usepackage[hidelinks]{hyperref}

\setlist[itemize]{leftmargin=*,nosep}
\setlist[enumerate]{leftmargin=*,nosep}
\newcommand{\dataset}[1]{\textsc{#1}}
\newcommand{\model}[1]{\textsc{#1}}

\title{Islamic Large Language Models: From Knowledge Acquisition to Trustworthy and Hallucination-Resistant AI}
\author{Mohammed Amine Mouhoub \\
Paris Dauphine University \\
\texttt{mohamed.mouhoub@dauphine.fr}}
\date{}

\begin{document}
\maketitle

\begin{abstract}
Large language models (LLMs) are increasingly used for knowledge-intensive question answering, including religious and legal questions. Islamic knowledge is a particularly demanding setting: answers are expected to be grounded in authoritative sources, citations must be exact, Arabic varieties differ substantially from the language of classical sources, and legitimate jurisprudential disagreement must be represented rather than collapsed into a single answer. This survey reviews the emerging field of Islamic LLMs and trustworthy Islamic AI. We organize the literature around Arabic NLP and Arabic-centric LLMs, Islamic NLP resources, Qur'anic question answering, Islamic knowledge benchmarks, retrieval-augmented generation, Islamic legal reasoning, inheritance reasoning, hallucination evaluation, and trustworthiness. We argue that fluency in Arabic is not sufficient for Islamic AI. Reliable systems require curated sources, retrieval and verification modules, citation-aware generation, madhhab-aware reasoning, human expert evaluation, and benchmarks that measure not only answer accuracy but also faithfulness, source validity, and reasoning quality. The survey concludes with a research agenda for hallucination-resistant Islamic AI systems.
\end{abstract}

\section{Introduction}

Large language models have transformed natural language processing by enabling strong performance in text generation, question answering, summarization, dialogue, and reasoning. General-purpose LLMs inherit many capabilities from large-scale pretraining, but they also inherit weaknesses: they may reproduce false beliefs, generate unsupported claims, fabricate citations, and appear confident even when their answers are not grounded in evidence. These limitations are visible in general factuality benchmarks such as \dataset{TruthfulQA}, where scaling alone did not guarantee truthfulness \citep{lin2022truthfulqa}, and in hallucination-oriented evaluations such as \dataset{RAGTruth} and \dataset{Mu-SHROOM}, where errors remain observable even when models are conditioned on retrieved evidence \citep{niu2024ragtruth,vazquez2025mushroom}.

Islamic knowledge is a high-stakes domain in which these problems become especially important. A wrong date in a historical answer may be inconvenient; a fabricated Hadith citation, a wrong Qur'anic verse number, or an unsupported Fatwa-style ruling can mislead users about religious practice. Islamic scholarship is also structurally different from many open-domain knowledge settings. It is grounded in the Qur'an, Hadith collections, Tafsir, Fiqh, legal opinions, historical scholarship, and a long tradition of juristic reasoning. Answers often depend on source hierarchy, authenticity, context, linguistic interpretation, and madhhab-specific methodology. A model that merely generates a fluent answer in Arabic does not necessarily satisfy these requirements.

The field has recently moved from Arabic language modeling toward Islamic-specific evaluation. Arabic pretrained language models such as \model{AraBERT}, \model{ARBERT}, \model{MARBERT}, and \model{CAMeLBERT} improved Arabic understanding by modeling Modern Standard Arabic, dialectal Arabic, and Classical Arabic varieties \citep{antoun2020arabert,abdulmageed2021arbert,inoue2021camelbert}. Arabic-centric LLMs such as \model{Jais}, \model{ALLaM}, \model{AceGPT}, \model{Fanar}, and \model{AraLLaMA} extended these advances to open-ended generation and instruction following \citep{sengupta2023jais,bari2024allam,huang2024acegpt,fanar2025,zhu2025arallama}. In parallel, Islamic benchmarks such as \dataset{Qur'an QA}, \dataset{IslamicMMLU}, \dataset{IslamicEval}, \dataset{QIAS}, and \dataset{MAWARITH} have provided task-specific evaluation for Qur'anic QA, Islamic knowledge, hallucination detection, and Islamic legal reasoning \citep{malhas2022quranqa,malhas2023quranqa,abdelaal2026islamicmmlu,mubarak2025islamiceval,bouchekif2025qias,bouchekif2026qias,bouchekif2026mawarith}.

This survey examines this transition from Arabic NLP to trustworthy Islamic AI. The central argument is that Islamic LLMs should be evaluated and designed around evidence, not only around language fluency or benchmark accuracy. A reliable Islamic assistant must retrieve authoritative sources, verify citations, distinguish direct textual evidence from juristic inference, acknowledge disagreement, and avoid overconfident Fatwa-like outputs. These requirements place Islamic AI at the intersection of Arabic NLP, retrieval-augmented generation, legal reasoning, factuality evaluation, and human-centered governance.

The contributions of this survey are fourfold. First, we synthesize the evolution of Arabic language technology from morphological tools and dialect identification to Arabic-centric LLMs. Second, we review Islamic datasets and benchmarks, with emphasis on the capabilities they measure and the gaps they leave unresolved. Third, we analyze hallucination failure modes that are specific to Islamic content and connect them to broader hallucination research. Fourth, we propose a trustworthiness framework for Islamic AI based on source grounding, citation verification, madhhab awareness, hallucination control, and scholar oversight.

\section{Foundations of Islamic Knowledge}

Islamic knowledge is not a flat collection of facts. It is an evidence-based tradition organized around primary sources, secondary interpretation, legal methodology, and scholarly disagreement. This structure has direct consequences for language modeling. A model must not only retrieve relevant text but also understand which source has authority, whether a statement is a direct quotation or an interpretation, and whether a ruling is universal, school-specific, contested, or context-dependent.

The Qur'an is the central source of Islamic knowledge. Computationally, it is challenging because Qur'anic Arabic differs from Modern Standard Arabic in vocabulary, morphology, orthography, rhetoric, and discourse structure. The Qur'anic Arabic Corpus introduced detailed morphological annotation for Quranic Arabic and emphasized that Quranic language cannot simply be treated as ordinary modern Arabic text \citep{dukes2010morphological}. For LLMs, this matters because surface similarity can be misleading: a question may use contemporary wording while the relevant verse expresses the concept through classical phrasing or indirect narrative structure.

Hadith literature adds another layer of complexity. A Hadith normally includes a chain of transmission and a textual content segment, and its evidential status depends on authenticity judgments, narrator reliability, and scholarly classification. A model that retrieves the text of a narration but ignores its authenticity grade may produce a misleading answer. Conversely, a model that fabricates a plausible Hadith phrase can create a particularly serious hallucination because users may treat it as prophetic evidence. This makes Hadith QA different from ordinary document QA: source metadata and authenticity are part of the answer.

Secondary sources include Tafsir, Fiqh, Fatwas, Seerah, and Islamic history. Tafsir explains Qur'anic verses using linguistic analysis, historical context, Hadith evidence, and scholarly interpretation. Fiqh applies legal methodology to practical questions and is shaped by multiple schools of law. Fatwas are context-sensitive legal opinions issued by qualified scholars. Seerah and Islamic history involve temporal and biographical reasoning. These source types require different evaluation criteria. Exact-match QA metrics may be useful for extracting a short answer from a passage, but they are insufficient for evaluating whether a Fatwa-style response is jurisprudentially valid.

Jurisprudential diversity is a core challenge. Sunni legal traditions such as Hanafi, Maliki, Shafi'i, and Hanbali schools may disagree on practical questions while remaining within recognized Islamic jurisprudence. A trustworthy model should therefore avoid presenting a contested view as consensus. The challenge is not merely classification of schools; it is reasoning under explicit legal assumptions. A system answering a question about Zakat al-Fitr, prayer details, inheritance, or financial contracts should make clear whether the answer follows a particular school, a majority view, or a contemporary Fatwa body.

\begin{table}[t]
\centering
\small
\begin{tabularx}{\linewidth}{lX}
\toprule
\textbf{Requirement} & \textbf{Consequence for Islamic AI}\\
\midrule
Source hierarchy & Distinguish Qur'an, Hadith, Tafsir, Fiqh, Fatwa, and historical reports.\\
Citation precision & Verify verse, collection, narrator, book/chapter, and authenticity where applicable.\\
Classical Arabic & Bridge user queries in MSA or dialects to classical source language.\\
Juristic disagreement & Represent madhhab-specific and contested positions explicitly.\\
Legal reasoning & Apply rules and exceptions rather than only retrieve passages.\\
Safety boundary & Avoid autonomous Fatwa issuance in high-stakes cases.\\
\bottomrule
\end{tabularx}
\caption{Requirements that distinguish Islamic AI from ordinary open-domain QA.}
\label{tab:requirements}
\end{table}

\section{From Arabic NLP to Arabic LLMs}

Islamic LLMs are built on decades of Arabic NLP. Arabic morphology, cliticization, diacritics, spelling variation, and dialectal diversity made Arabic NLP challenging before the transformer era. Tools such as \model{MADAMIRA} and \model{Farasa} addressed core preprocessing needs. \model{MADAMIRA} combined morphological analysis and disambiguation in a robust Arabic processing system \citep{pasha2014madamira}, while \model{Farasa} provided a fast Arabic segmenter used in downstream tasks such as machine translation and information retrieval \citep{abdelali2016farasa}. \model{CAMeL Tools} later offered an open-source Python toolkit for preprocessing, morphological modeling, dialect identification, named entity recognition, and sentiment analysis \citep{obeid2020cameltools}. These resources matter for Islamic AI because source normalization and morphological analysis remain important when aligning user queries with classical texts.

Arabic dialect identification was another important precursor. Many users ask religious questions in dialectal Arabic, while authoritative sources are in Classical Arabic or MSA. \citet{defrancony2019hierarchical} proposed hierarchical deep learning for Arabic dialect identification, exploiting dialect relationships rather than treating all dialect labels as unrelated. This line of work remains relevant because a practical Islamic assistant must handle dialectal questions, normalize them, and retrieve evidence from formal or classical corpora.

The transformer era produced Arabic pretrained language models. \model{AraBERT} showed that Arabic-specific BERT pretraining improves performance on Arabic understanding tasks compared with multilingual baselines \citep{antoun2020arabert}. \model{ARBERT} and \model{MARBERT} extended the ecosystem to diverse Arabic varieties; MARBERT was particularly important for dialectal and social-media Arabic \citep{abdulmageed2021arbert}. \model{CAMeLBERT} explicitly studied the relationship between Arabic variant, data size, and downstream task type, and released variants for MSA, dialectal Arabic, Classical Arabic, and mixed Arabic \citep{inoue2021camelbert}. The Classical Arabic variant is especially relevant for Islamic texts.

Arabic-centric generative LLMs shifted the focus from representation learning to instruction following and open-ended generation. \model{Jais} introduced Arabic-centric foundation and instruction-tuned generative models based on a GPT-style architecture and trained on Arabic and English data \citep{sengupta2023jais}. \model{ALLaM} explored vocabulary expansion, Arabic-English pretraining, and alignment for Arabic and English capabilities \citep{bari2024allam}. \model{AceGPT} focused on localization and cultural alignment in Arabic, highlighting that linguistic competence alone is not enough when models interact with culturally specific user expectations \citep{huang2024acegpt}. \model{Fanar} presented an Arabic-centric multimodal generative AI platform with Fanar Star and Fanar Prime, and notably reported an Islamic RAG capability and attribution service as part of the platform \citep{fanar2025}. \model{AraLLaMA} studied progressive vocabulary expansion as a route for Arabic acquisition in LLMs \citep{zhu2025arallama}.

The key lesson for Islamic AI is that Arabic fluency and Islamic reliability are separable. A model may write elegant Arabic while hallucinating citations or confusing legal schools. Conversely, a smaller model with retrieval, citation verification, and rule-based tools may outperform a larger model on constrained religious tasks. This explains why the field is moving from model-centric evaluation toward system-level evaluation that includes retrieval, verification, and specialized reasoning modules.

\begin{table*}[t]
\centering
\small
\begin{tabularx}{\textwidth}{l l l X X}
\toprule
\textbf{Resource} & \textbf{Period} & \textbf{Type} & \textbf{Main Contribution} & \textbf{Relevance to Islamic AI}\\
\midrule
MADAMIRA & 2014 & Morphology & Arabic morphological analysis and disambiguation. & Normalization and analysis of complex Arabic forms.\\
Farasa & 2016 & Segmenter & Fast Arabic segmentation. & Query and source preprocessing.\\
CAMeL Tools & 2020 & Toolkit & Open-source Arabic NLP utilities. & Practical preprocessing and dialect-aware pipelines.\\
Hierarchical dialect ID & 2019 & Neural classifier & Hierarchical modeling of Arabic dialects. & Bridges dialectal user queries and formal sources.\\
AraBERT & 2020 & PLM & Arabic BERT-style pretraining. & Baseline encoder for Arabic understanding.\\
ARBERT/MARBERT & 2021 & PLM & Models for MSA and dialectal Arabic. & Improves robustness to user language variation.\\
CAMeLBERT & 2021 & PLM & MSA, dialectal, Classical, and mixed variants. & Classical Arabic coverage for Islamic texts.\\
Jais & 2023 & LLM & Arabic-centric generative LLM. & Baseline for Arabic instruction following.\\
ALLaM & 2024 & LLM & Arabic-English LLM with vocabulary expansion and alignment. & Arabic knowledge and bilingual Islamic QA.\\
AceGPT & 2024 & LLM & Arabic localization and cultural alignment. & Highlights culture-sensitive evaluation.\\
Fanar & 2025 & Platform & Arabic multimodal platform with Islamic RAG. & Islamic QA and attribution-oriented generation.\\
AraLLaMA & 2025 & LLM & Progressive vocabulary expansion. & Efficient Arabic acquisition strategy.\\
\bottomrule
\end{tabularx}
\caption{Representative milestones from Arabic NLP to Arabic-centric LLMs.}
\label{tab:arabic_models}
\end{table*}

\section{Islamic NLP Before and During the LLM Era}

Islamic NLP began before the current LLM wave. Early work focused on Quranic annotation, verse retrieval, reading comprehension, Hadith and Fatwa information access, and Islamic knowledge organization. These lines of work remain important because LLM-based systems still require curated corpora, retrieval indices, and expert-verified evaluation sets. In other words, LLMs do not replace Islamic NLP resources; they increase the need for them.

Qur'anic NLP has been the most visible subfield. The Qur'anic Arabic Corpus is a foundational resource because it provides morphological and syntactic analysis aligned with traditional Arabic grammar \citep{dukes2010morphological}. Qur'an QA 2022 organized the first shared task on question answering over the Holy Qur'an and framed it as machine reading comprehension \citep{malhas2022quranqa}. Qur'an QA 2023 extended this direction by separating passage retrieval and machine reading comprehension, attracting multiple systems and highlighting the importance of retrieval quality before answer extraction \citep{malhas2023quranqa}. This progression mirrors a broader pattern: early systems emphasized extracting answers, whereas newer systems increasingly evaluate retrieval, grounding, and evidence selection.

Hadith processing is less standardized in benchmark form, but it is central for Islamic AI. A Hadith answer may require exact text, collection, narrator, authenticity, and relationship to other narrations. This makes Hadith QA more metadata-dependent than many open-domain QA tasks. A model may retrieve a relevant narration but misclassify its authenticity or cite the wrong collection. In LLM settings, the risk is more severe: the model may fabricate a plausible narration, a source title, or a chain of transmission. IslamicEval explicitly targets hallucination in Qur'anic and Hadith content, reflecting the importance of this failure mode \citep{mubarak2025islamiceval}.

Islamic legal and Fatwa-oriented NLP is more difficult because the target is not merely a fact but a judgment. Legal answers involve evidence, methodology, and context. The emergence of Islamic legal reasoning benchmarks such as QIAS 2025, QIAS 2026, and MAWARITH therefore marks a shift from retrieval and reading comprehension toward reasoning evaluation \citep{bouchekif2025qias,bouchekif2026qias,bouchekif2026mawarith}. The recently proposed IslamicLegalBench also reflects the growing need to evaluate pluralist Islamic legal traditions across multiple schools, not only single-answer knowledge recall \citep{elmahjub2026islamiclegalbench}.

This history suggests that Islamic LLM research should not be framed as a replacement for earlier Islamic NLP. Instead, it should be viewed as a system-building layer above earlier resources. Strong systems will combine annotated corpora, retrieval, symbolic constraints, expert-curated benchmarks, and LLM generation. Weak systems will simply prompt general LLMs and rely on fluency.

\section{Datasets and Benchmarks}

A useful benchmark does more than assign a score; it defines what the community values. Early Islamic benchmarks measured reading comprehension and passage retrieval. Recent benchmarks increasingly measure knowledge coverage, legal reasoning, hallucination detection, and source faithfulness. Table~\ref{tab:benchmarks} summarizes the verified resources that currently anchor the field.

\begin{table*}[t]
\centering
\small
\begin{tabularx}{\textwidth}{l l l l X}
\toprule
\textbf{Benchmark} & \textbf{Year} & \textbf{Domain} & \textbf{Task} & \textbf{What it Measures}\\
\midrule
Qur'an QA 2022 & 2022 & Qur'an & MRC & Extractive QA over Qur'anic passages.\\
Qur'an QA 2023 & 2023 & Qur'an & PR + MRC & Retrieval and reading comprehension over the Qur'an.\\
IslamicEval & 2025 & Islamic content & Hallucination & Identification, validation, correction, and relevance of Islamic content hallucinations.\\
QIAS 2025 & 2025 & Inheritance and knowledge & Shared task & Religious and legal reasoning with expert-validated MCQs.\\
QIAS 2026 & 2026 & Inheritance & Shared task & End-to-end Arabic inheritance reasoning with structured traces and MIR-E multi-step evaluation.\\
Inheritance Law Eval. & 2025 & Inheritance & MCQ reasoning & Knowledge and reasoning gaps in Islamic inheritance law.\\
IslamicMMLU & 2026 & Qur'an/Hadith/Fiqh & MCQ & 10,013 questions across three Islamic knowledge tracks, including madhhab bias analysis.\\
MAWARITH & 2026 & Inheritance & Full-chain reasoning & 12,500 inheritance cases with step-by-step legal and numerical reasoning.\\
Fanar-Sadiq Eval. & 2026 & Islamic QA & System evaluation & Agentic Islamic QA with citation normalization, verification traces, and calculators.\\
IslamicLegalBench & 2026 & Islamic law & Legal benchmark & Islamic legal knowledge and reasoning across pluralist traditions.\\
PalmX 2025 & 2025 & Arabic/Islamic culture & Shared task & Cultural and Islamic knowledge competence in MCQ format.\\
\bottomrule
\end{tabularx}
\caption{Verified Islamic and Arabic-Islamic benchmarks discussed in this survey. PR: passage retrieval; MRC: machine reading comprehension.}
\label{tab:benchmarks}
\end{table*}

The Qur'an QA shared tasks are important because they provided a public evaluation environment for Qur'anic QA. Qur'an QA 2022 focused on machine reading comprehension, whereas Qur'an QA 2023 included both passage retrieval and MRC. This distinction is important: if a system fails to retrieve the right passage, even a strong reader cannot answer correctly. The 2023 task therefore moved closer to real-world Islamic QA, where retrieval is often the bottleneck.

IslamicMMLU broadens evaluation from Qur'anic QA to Islamic knowledge. It contains 10,013 multiple-choice questions across Qur'an, Hadith, and Fiqh tracks and evaluates 26 LLMs, with reported accuracy varying widely across models \citep{abdelaal2026islamicmmlu}. Its inclusion of madhhab bias detection is especially significant because it recognizes that Islamic legal evaluation cannot always rely on a single universal answer. Nevertheless, multiple-choice evaluation remains limited: it does not measure citation accuracy, reasoning faithfulness, or the quality of generated explanations.

IslamicEval targets a different problem: hallucination in Islamic content. Its design reflects the observation that LLMs may produce plausible but unsupported religious text. The shared task includes complementary subtasks for identifying Islamic content, validating it, correcting hallucinations, and checking relevance \citep{mubarak2025islamiceval}. This is a critical development because hallucination evaluation in religious settings needs specialized criteria. A fabricated source is not only a factual error; it is a failure of epistemic responsibility.

QIAS and MAWARITH introduce reasoning-centric evaluation. QIAS 2025 was organized as a shared task on Islamic inheritance reasoning and Islamic knowledge assessment \citep{bouchekif2025qias}. MAWARITH then moved beyond multiple-choice inheritance questions by providing 12,500 Arabic inheritance cases with full reasoning chains and exact share computations \citep{bouchekif2026mawarith}. QIAS 2026 further operationalized this setting as an end-to-end shared task: systems were required to read Arabic inheritance cases, generate structured reasoning traces, identify eligible and blocked heirs, assign shares, handle adjustment cases, and compute final distributions under the MIR-E multi-stage metric \citep{bouchekif2026qias}. This progression is important because many LLMs can produce convincing explanations while making intermediate legal or arithmetic mistakes. Stage-wise evaluation is therefore more informative than final-answer accuracy.

\section{Islamic Question Answering}

Islamic QA can be organized into retrieval-based QA, neural reading comprehension, LLM-based QA, and retrieval-augmented generation. The earliest retrieval systems prioritized source access: given a question, they returned relevant verses, Hadiths, or Fatwa passages. Such systems are often more transparent than LLMs because their output is directly tied to retrieved text. However, they struggle with paraphrase, multi-hop questions, and synthesis across sources.

Neural QA improved semantic matching and extraction. Qur'an QA tasks, for example, encouraged models to identify answer spans in Qur'anic passages \citep{malhas2022quranqa,malhas2023quranqa}. Yet extractive QA is still limited because many Islamic questions require interpretation. A user may ask for the ruling on a practical issue; the answer may require a verse, a Hadith, a principle of Fiqh, and a school-specific interpretation. Extracting a span from one passage does not solve the problem.

LLM-based QA offers natural explanations and dialogue but introduces hallucination. A fluent answer can hide unsupported claims. This is why Islamic QA should be evaluated not only by answer quality but also by evidence quality. The most promising systems combine LLM generation with retrieval, citation normalization, and verification, as illustrated by Fanar-Sadiq \citep{abbas2026fanarsadiq}. Such architectures treat the LLM as one component in a larger system rather than as an autonomous religious authority.

Different Islamic QA domains require different constraints. Qur'anic QA requires exact verse retrieval and careful interpretation. Hadith QA requires authenticity metadata and source attribution. Fatwa QA requires context and juristic methodology. Inheritance QA requires rule application and arithmetic. Open-domain Islamic QA requires intent classification because the user's question may demand quotation, explanation, legal reasoning, or refusal. A single prompt template cannot handle all of these cases reliably.

\begin{table}[t]
\centering
\small
\begin{tabularx}{\linewidth}{lX}
\toprule
\textbf{QA Type} & \textbf{Main Risk}\\
\midrule
Qur'anic QA & Wrong verse retrieval or over-interpretation without Tafsir.\\
Hadith QA & Fabricated narration or wrong authenticity attribution.\\
Fatwa QA & Unsupported ruling or missing context.\\
Inheritance QA & Incorrect heir identification, rule application, or arithmetic.\\
Historical QA & Wrong dates, names, or event relations.\\
Open-domain QA & Intent mismatch and overconfident synthesis.\\
\bottomrule
\end{tabularx}
\caption{Islamic QA types and characteristic risks.}
\label{tab:qa_risks}
\end{table}

\section{Retrieval-Augmented and Agentic Islamic AI}

Retrieval-augmented generation is often presented as a mitigation for hallucination, but in Islamic AI it is more than mitigation: it is a design requirement. Islamic answers are expected to be evidence-based. The relevant evidence may come from the Qur'an, Hadith, Tafsir, Fiqh manuals, Fatwa repositories, or historical sources. Retrieval therefore supplies not only facts but also epistemic legitimacy.

Classical RAG retrieves documents and conditions a generator on them \citep{lewis2020rag}. Sparse methods such as BM25 remain useful for exact lexical matching and citation retrieval \citep{robertson2009bm25}, while dense retrieval improves semantic matching \citep{karpukhin2020dpr}. Hybrid retrieval is often attractive in Islamic QA because both lexical precision and semantic recall matter. A verse or Hadith reference may require exact wording, while a user question may be phrased indirectly or dialectally. General retrieval benchmarks such as BEIR show the difficulty of robust retrieval across domains \citep{thakur2021beir}; Islamic retrieval adds source hierarchy and citation granularity.

However, standard retrieve-then-generate pipelines are insufficient. RAG systems can still hallucinate by misusing retrieved evidence, adding unsupported claims, or citing irrelevant passages. RAGTruth demonstrates that word-level hallucinations remain present in RAG settings and that manual fine-grained annotation is necessary to study them \citep{niu2024ragtruth}. In Islamic AI, these failures appear as citation drift, evidence overreach, and legal overgeneralization. A model might retrieve a verse about one context and use it to justify a ruling in another without explaining the juristic bridge.

Agentic Islamic AI attempts to address these limitations by decomposing the task. Fanar-Sadiq routes Islamic queries to specialized modules and supports intent-aware routing, retrieval-grounded Fiqh answers, citation normalization, verification traces, exact verse lookup, and deterministic calculators for zakat and inheritance \citep{abbas2026fanarsadiq}. This design is important because it recognizes that religious questions are heterogeneous. Some require exact quotation, some require legal reasoning, and some require deterministic computation. A monolithic LLM response is not sufficient for all of them.

\begin{figure}[t]
\centering
\begin{tikzpicture}[node distance=0.52cm, box/.style={rectangle,rounded corners,draw=black,align=center,font=\small,minimum width=6.1cm,minimum height=0.55cm}, arrow/.style={-Latex,thick}]
\node[box] (q) {User question};
\node[box, below=of q] (intent) {Intent classifier: quote, QA, Fatwa, calculator, history};
\node[box, below=of intent] (retr) {Hybrid retrieval over curated Islamic sources};
\node[box, below=of retr] (reason) {Reasoning module: evidence synthesis or rule-based tool};
\node[box, below=of reason] (verify) {Verification: citation, quotation, faithfulness, arithmetic};
\node[box, below=of verify] (ans) {Cautious answer with sources and uncertainty};
\draw[arrow] (q) -- (intent);
\draw[arrow] (intent) -- (retr);
\draw[arrow] (retr) -- (reason);
\draw[arrow] (reason) -- (verify);
\draw[arrow] (verify) -- (ans);
\end{tikzpicture}
\caption{A retrieval-verification-generation architecture for trustworthy Islamic QA.}
\label{fig:rag_framework}
\end{figure}
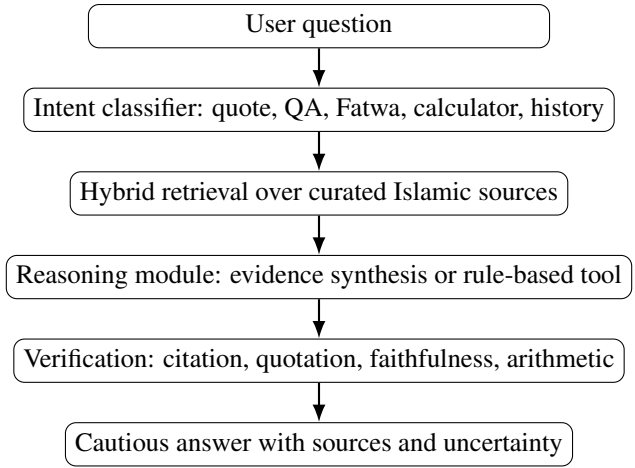

The next generation of Islamic RAG systems should evaluate not only retrieval relevance but also evidence authority. A retrieved blog post and a primary Hadith collection do not carry the same weight. Similarly, retrieving a chapter when the answer needs a precise verse is inadequate. Authority-aware and granularity-aware retrieval metrics are therefore needed. Systems should also track whether generated claims are entailed by the evidence, whether citations are normalized, and whether conflicting evidence has been addressed.

\section{Islamic Legal Reasoning}

Islamic legal reasoning is a stronger test of model competence than factual knowledge. Legal questions require identifying relevant evidence, applying methodology, handling exceptions, and acknowledging disagreement. General legal LLM benchmarks already show that legal reasoning cannot be reduced to ordinary QA. Islamic law adds further complexity because jurisprudential schools may differ in their principles, terminology, and practical rulings.

Inheritance law is an especially useful benchmark because it combines legal and numerical reasoning. Solving an inheritance case requires understanding the family scenario, identifying eligible heirs, applying blocking rules, assigning fixed shares, handling residual distributions, and computing exact fractions. This structure makes error propagation visible. A model that misidentifies one heir may produce a plausible but entirely wrong distribution.

The 2025 inheritance evaluation by \citet{bouchekif2025assessing} tested seven LLMs on 1,000 multiple-choice inheritance questions and found a substantial gap between stronger reasoning models and Arabic or open models. QIAS 2025 expanded the evaluation into a shared task setting for Islamic inheritance reasoning and Islamic knowledge assessment \citep{bouchekif2025qias}. MAWARITH then moved from multiple-choice evaluation to full-chain reasoning, providing 12,500 Arabic cases with intermediate legal decisions and exact share calculations \citep{bouchekif2026mawarith}. QIAS 2026 used MAWARITH as the basis for a more demanding end-to-end shared task, with 12,000 training cases and 500 test cases distributed across simple, \emph{`awl}, and \emph{radd} categories, and evaluated submissions with MIR-E, a multi-step metric covering heirs and blocking, share assignment, adjustment prediction, and final distribution \citep{bouchekif2026qias}. This progression is important because multiple-choice answers can overestimate competence: a model may select the correct option without producing valid reasoning.

Islamic legal reasoning also requires madhhab awareness. IslamicMMLU includes a Fiqh track and a madhhab bias analysis task \citep{abdelaal2026islamicmmlu}. IslamicLegalBench explicitly focuses on Islamic law across pluralist legal traditions and reports limitations in LLM correctness and hallucination \citep{elmahjub2026islamiclegalbench}. These efforts indicate that the field is beginning to move beyond generic Islamic knowledge toward jurisprudentially structured evaluation.

\begin{table}[t]
\centering
\small
\begin{tabularx}{\linewidth}{lX}
\toprule
\textbf{Failure Mode} & \textbf{Description}\\
\midrule
Scenario error & Misreads family relations or relevant facts.\\
Heir error & Includes excluded heirs or excludes eligible heirs.\\
Rule error & Applies wrong fixed share, blocking rule, awl, or radd.\\
Arithmetic error & Correct legal reasoning but wrong numerical computation.\\
Madhhab error & Mixes rules from different legal schools.\\
Explanation error & Provides plausible but legally invalid justification.\\
\bottomrule
\end{tabularx}
\caption{Frequent error types in Islamic inheritance and legal reasoning.}
\label{tab:legal_errors}
\end{table}

The design implication is clear: Islamic legal AI should not be evaluated only by final answers. It should be evaluated through staged reasoning, evidence links, legal assumptions, and numerical consistency. Neuro-symbolic systems may be valuable here. LLMs can parse scenarios and explain reasoning, while symbolic calculators or rule engines can enforce legal invariants. Fanar-Sadiq's use of deterministic calculators for zakat and inheritance is an example of this direction \citep{abbas2026fanarsadiq}.

The official QIAS 2026 leaderboard also provides useful methodological evidence. The top systems were not simple unconstrained prompting baselines: CVPD obtained the highest reported MIR-E score with a hybrid RAG pipeline, Simplicity used a two-stage neuro-symbolic design that separated Arabic information extraction from legal computation, and QU-NLP showed that multi-stage QLoRA adaptation can make smaller open-weight models competitive. These results suggest that inheritance reasoning benefits from structured generation, retrieval or domain adaptation, and explicit validation constraints rather than relying on raw generative fluency alone \citep{bouchekif2026qias}.

\section{Hallucinations in Islamic LLMs}

Hallucination is the central trustworthiness problem for Islamic LLMs. General hallucination research distinguishes false answers, unsupported generation, and overgeneration. TruthfulQA showed that models can imitate common falsehoods \citep{lin2022truthfulqa}. HaluEval studied hallucination in general LLM outputs \citep{li2023halueval}. SelfCheckGPT proposed a black-box method based on consistency across sampled responses \citep{manakul2023selfcheckgpt}. FActScore evaluated factual precision using atomic facts \citep{min2023factscore}. RAGTruth and Mu-SHROOM moved toward fine-grained and span-level hallucination evaluation \citep{niu2024ragtruth,vazquez2025mushroom}. Islamic hallucination research should build on these methods but adapt them to religious epistemology.

Islamic hallucinations can be organized by the level at which the error occurs. Source-level hallucinations fabricate or misquote primary sources. Citation-level hallucinations cite the wrong Surah, verse, Hadith collection, narrator, or numbering. Attribution-level hallucinations assign a view to the wrong scholar or madhhab. Legal hallucinations produce unsupported rulings. Reasoning hallucinations apply invalid inferences from valid evidence. Historical hallucinations invent events, dates, or biographies. Madhhab hallucinations merge incompatible school-specific rulings or present a contested opinion as consensus.

\begin{table*}[t]
\centering
\small
\begin{tabularx}{\textwidth}{l l X X}
\toprule
\textbf{Level} & \textbf{Type} & \textbf{Typical Example} & \textbf{Preferred Mitigation}\\
\midrule
Source & Fabricated Hadith & Invented narration attributed to a canonical collection. & Hadith database retrieval and quotation validation.\\
Citation & Wrong verse or reference & Correct theme but wrong Surah/ayah number. & Exact citation lookup and normalization.\\
Attribution & Wrong scholar or madhhab & Opinion attributed to the wrong legal school. & Source-linked scholar and school metadata.\\
Legal & Unsupported ruling & Fatwa-style answer without evidence or conditions. & Evidence-grounded generation and refusal policies.\\
Reasoning & Invalid inference & Evidence is real but applied to an unrelated case. & NLI-style faithfulness and expert review.\\
Madhhab & School confusion & Hanafi and Shafi'i rulings merged as consensus. & Madhhab-aware routing and explicit assumptions.\\
Historical & False event or date & Plausible but incorrect historical claim. & Chronology-aware retrieval and verification.\\
\bottomrule
\end{tabularx}
\caption{A taxonomy of hallucinations in Islamic LLMs.}
\label{tab:halluc_taxonomy}
\end{table*}

IslamicEval is a major step because it focuses directly on hallucination in Islamic content and includes subtasks for validation and correction \citep{mubarak2025islamiceval}. The HUMAIN submission to IslamicEval emphasizes that even small deviations in Qur'anic or Hadith content are unacceptable in religious settings \citep{omayrah2025humain}. This differs from many open-domain contexts where approximate paraphrases may be tolerated. For Qur'anic text, exact quotation matters; for Hadith, authenticity and source matter.

Fine-grained evaluation is particularly important. A long answer may be mostly correct but contain one fabricated reference. Binary labels treat the whole answer as hallucinated or not, whereas span-level localization identifies the exact unsupported segment. This is why RAGTruth and Mu-SHROOM are methodologically relevant for Islamic AI \citep{niu2024ragtruth,vazquez2025mushroom}. Islamic datasets should similarly annotate spans, evidence links, and correction targets.

Detection methods fall into several categories. Retrieval-based verification compares generated claims against authoritative sources. LLM-as-judge methods can identify likely unsupported claims but must themselves be constrained and audited. Self-consistency methods can detect unstable claims, but they are weaker for entrenched false beliefs. Natural language inference can test whether evidence entails a claim, but Islamic legal reasoning often requires more than textual entailment. Human expert evaluation remains essential for Fatwa-like and jurisprudential claims.

Correction is harder than detection. A system must remove unsupported content, retrieve valid evidence, reformulate the answer, and preserve uncertainty. If no evidence is found, the correct correction may be refusal or a statement of insufficient evidence. In Islamic AI, it is better to say that the available sources do not support a claim than to generate a plausible substitute.

\section{Toward Trustworthy Islamic AI}

Trustworthy Islamic AI should be designed around five pillars: source grounding, citation verification, madhhab awareness, hallucination control, and human scholar oversight. These pillars are mutually reinforcing. Source grounding without citation verification may still produce wrong references. Citation verification without madhhab awareness may still produce misleading legal answers. Automated hallucination detection without scholar oversight may miss subtle legal or theological errors.

Source grounding means that answers should be tied to authoritative evidence. For Qur'anic answers, this includes exact Surah and verse references. For Hadith answers, it includes collection, narrator, authenticity where relevant, and exact quotation. For Fiqh answers, it includes legal school, source text, and whether the answer is a majority, minority, or contemporary view. Grounding should be visible to users and auditable by developers.

Citation verification should be treated as a separate module. LLMs are weak at exact reference generation because they optimize plausible continuation, not database consistency. A reliable system should verify that every cited verse or Hadith exists and that it supports the claim. Fanar-Sadiq's citation normalization and verification traces illustrate this architectural direction \citep{abbas2026fanarsadiq}.

Madhhab awareness requires explicit legal assumptions. A system should ask clarifying questions when the school is relevant, or answer by presenting multiple views. It should not silently merge school-specific rulings. Benchmarks such as IslamicMMLU and IslamicLegalBench show that madhhab-sensitive evaluation is now becoming an explicit research topic \citep{abdelaal2026islamicmmlu,elmahjub2026islamiclegalbench}.

Human scholar oversight is not optional for high-stakes religious guidance. Scholars should participate in dataset design, annotation guidelines, error analysis, benchmark validation, and system governance. LLMs can assist by retrieving sources, summarizing opinions, and drafting explanations, but autonomous Fatwa issuance remains risky. A trustworthy system should make this boundary clear to users.

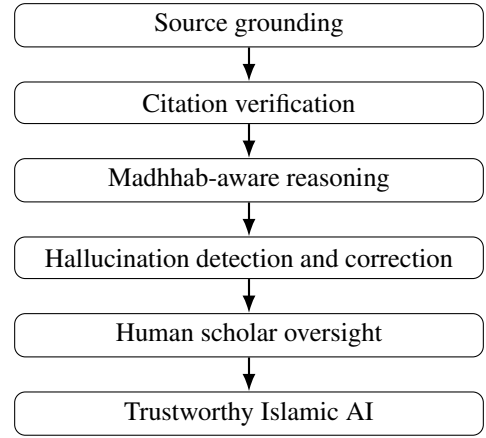
\begin{figure}[t]
\centering
\begin{tikzpicture}[node distance=0.45cm, box/.style={rectangle,rounded corners,draw=black,align=center,font=\small,minimum width=6.2cm,minimum height=0.55cm}, arrow/.style={-Latex,thick}]
\node[box] (ground) {Source grounding};
\node[box, below=of ground] (cite) {Citation verification};
\node[box, below=of cite] (madhhab) {Madhhab-aware reasoning};
\node[box, below=of madhhab] (hall) {Hallucination detection and correction};
\node[box, below=of hall] (human) {Human scholar oversight};
\node[box, below=of human] (trust) {Trustworthy Islamic AI};
\draw[arrow] (ground) -- (cite);
\draw[arrow] (cite) -- (madhhab);
\draw[arrow] (madhhab) -- (hall);
\draw[arrow] (hall) -- (human);
\draw[arrow] (human) -- (trust);
\end{tikzpicture}
\caption{Five-pillar framework for trustworthy Islamic AI.}
\label{fig:trust_framework}
\end{figure}

\section{Evaluation Methodologies}

Evaluation of Islamic LLMs must be multidimensional. Multiple-choice accuracy is useful for broad knowledge benchmarks, but it does not measure citation accuracy, reasoning quality, or source faithfulness. IslamicMMLU provides a strong large-scale knowledge evaluation, but its MCQ format should be complemented by generative and evidence-grounded tasks \citep{abdelaal2026islamicmmlu}. Similarly, final-answer accuracy in inheritance problems should be complemented by stage-wise reasoning metrics, as argued by MAWARITH and operationalized in QIAS 2026 through MIR-E \citep{bouchekif2026mawarith,bouchekif2026qias}.

QIAS 2026 is a useful example of why reasoning evaluation needs to be decomposed. MIR-E separates heirs and blocking, share assignment, adjustment handling, and final distribution, making it possible to distinguish a legal interpretation error from a numerical calculation error \citep{bouchekif2026qias}. This kind of diagnostic metric is especially valuable in Islamic legal reasoning because errors propagate: a wrong heir set can make all subsequent shares incorrect even if the arithmetic procedure is otherwise coherent.

Retrieval evaluation should include Recall@k, MRR, and nDCG, but Islamic retrieval also needs authority-aware scoring. A system should be rewarded for retrieving primary sources when appropriate and for identifying exact verses or Hadiths rather than broad documents. In RAG settings, evaluation should measure whether generated claims are supported by retrieved evidence. RAGTruth demonstrates the value of word-level and response-level annotation for this purpose \citep{niu2024ragtruth}.

Hallucination evaluation should include detection precision, recall, F1, span-level F1, correction accuracy, and citation validity. Span-level evaluation is critical because Islamic hallucinations often occur in short but high-impact fragments: a fabricated source name, a wrong verse number, or an unsupported legal condition. Human expert evaluation should be used for legal and Fatwa-like answers because automatic metrics cannot fully capture jurisprudential validity.

\begin{table}[t]
\centering
\small
\begin{tabularx}{\linewidth}{lX}
\toprule
\textbf{Metric Family} & \textbf{Best Use}\\
\midrule
Accuracy & MCQ Islamic knowledge tasks.\\
Exact Match/F1 & Extractive Qur'anic reading comprehension.\\
Recall@k/MRR/nDCG & Retrieval over Qur'an, Hadith, Tafsir, and Fiqh.\\
Stage-wise reasoning & Inheritance and legal reasoning pipelines.\\
Span F1 & Hallucination localization.\\
Citation validity & Generated answers with references.\\
Expert rating & Fatwa quality and jurisprudential consistency.\\
\bottomrule
\end{tabularx}
\caption{Evaluation families for Islamic LLMs.}
\label{tab:metrics}
\end{table}

\section{Open Challenges}

The first open challenge is source coverage. Many Islamic texts are not available in clean, machine-readable, well-licensed formats. Even when texts are available, editions differ and metadata may be incomplete. This affects retrieval, citation normalization, and benchmark reproducibility. The second challenge is Arabic variation. User questions may be in dialectal Arabic, English, French, Urdu, or mixed language, while sources are often Classical Arabic. Systems must perform cross-variety and cross-lingual alignment without distorting meanings.

The third challenge is long-context and multi-source reasoning. Many Islamic questions require multiple pieces of evidence and interpretation across sources. Long context windows help but do not solve reasoning or source ranking. The fourth challenge is disagreement. Benchmarks must distinguish false answers from valid minority opinions. This is difficult because legal diversity is not noise; it is part of the domain.

The fifth challenge is hallucination under retrieval. RAG reduces hallucinations but does not eliminate them. Models may misuse retrieved evidence or generate unsupported legal bridges. The sixth challenge is expert annotation cost. High-quality Islamic datasets require scholars, linguists, and NLP experts. This makes data creation slower but also more valuable.

Finally, safety remains unresolved. Islamic assistants should not present themselves as Muftis. They should support learning, retrieval, and explanation while making boundaries clear. For sensitive cases, they should advise consulting qualified scholars.

\section{Future Research Directions}

Madhhab-aware LLMs are a natural next step. Instead of treating Islamic law as one homogeneous knowledge base, systems should represent school-specific evidence and reasoning patterns. This could involve school-conditioned retrieval, supervised fine-tuning on school-specific corpora, or mixture-of-experts architectures where different modules specialize in legal traditions.

Agentic Islamic AI is another promising direction. A robust assistant should classify intent, retrieve evidence, verify citations, reason over legal conditions, call deterministic calculators when necessary, and produce cautious answers. Such systems are more complex than simple chatbots, but Islamic QA requires this complexity. Fanar-Sadiq provides a strong example of agentic design for Islamic QA \citep{abbas2026fanarsadiq}.

Neuro-symbolic Fiqh systems are especially promising for structured domains such as inheritance and zakat. LLMs can parse user input and generate explanations, while symbolic modules enforce rules and arithmetic. MAWARITH shows why this is important: legal reasoning errors propagate across stages, and exact computations matter \citep{bouchekif2026mawarith}.

Future benchmarks should move beyond accuracy. They should evaluate faithfulness, citation validity, uncertainty, legal assumptions, and correction quality. Datasets should include span-level hallucination annotations, evidence links, and expert explanations. Public leaderboards should report not only a single score but also error categories that guide system improvement.

\section{Conclusion}

Islamic LLMs sit at the intersection of Arabic NLP, religious knowledge, legal reasoning, retrieval, and trustworthy AI. The field has advanced rapidly: Arabic PLMs and LLMs improved linguistic capability, Qur'an QA and IslamicMMLU provided knowledge benchmarks, IslamicEval focused attention on hallucination, and QIAS 2025, QIAS 2026, and MAWARITH introduced structured legal reasoning evaluation. Yet the core challenge remains: fluency is not reliability.

A trustworthy Islamic AI system must be evidence-grounded, citation-aware, jurisprudentially transparent, and cautious under uncertainty. It must distinguish primary sources from interpretation, recognize valid disagreement, verify citations, and avoid unsupported Fatwa-style generation. Future progress will depend less on prompting general models and more on building integrated systems with retrieval, verification, symbolic reasoning, expert data, and human oversight.

\section*{Limitations}
This manuscript is a complete survey draft, but it should undergo a final bibliographic audit before submission. Very recent 2026 works may change status, venue, or citation metadata. The final camera-ready version should replace this self-contained ACL-like format with the official ACL or TACL style files.

\end{document}